\documentclass[10pt,journal, compsoc]{IEEEtran}
\usepackage{amsmath,amsfonts}
\usepackage{array}
\usepackage{textcomp}
\usepackage{stfloats}
\usepackage{url}
\usepackage{verbatim}
\usepackage{graphicx}
\usepackage{cite}
\usepackage{dutchcal}
\usepackage{makecell}
\usepackage{threeparttable}
\usepackage[T1]{fontenc}
\usepackage{multirow}
\usepackage{booktabs}
\usepackage{blkarray}
\usepackage[linesnumbered,ruled,lined]{algorithm2e}
\usepackage{amsthm,amssymb}

\usepackage{mathrsfs}
\usepackage{subfigure}
\usepackage[hang]{footmisc}
\usepackage[font=small,labelfont=bf]{caption}
\usepackage[graphicx]{realboxes}
\usepackage{epstopdf}

\begin{document}
	
	\title{CONDEN-FI: Consistency and Diversity Learning-based Multi-View Unsupervised Feature and Instance Co-Selection}
	
	\author{Yanyong~Huang,~Yuxin~Cai,~Dongjie~Wang,~Xiuwen Yi,~and~Tianrui~Li,\IEEEmembership{Senior Member,~IEEE}
		\thanks{Yanyong~Huang and Yuxin~Cai are with the Joint Laboratory of Data Science and Business Intelligence, School of  Statistics,  Southwestern University of Finance and Economics, Chengdu 611130, China (e-mail: huangyy@swufe.edu.cn; caaaiyx@163.com);}
		\thanks{Dongjie~Wang is with the Department of Electrical Engineering and Computer Science, University of Kansas, Lawrence, KS 66045, USA (e-mail: wangdongjie@ku.edu);}
		\thanks{Xiuwen~Yi is with the JD Intelligent Cities Research and JD Intelligent Cities Business Unit, Beijing 100176, China (e-mail: xiuwenyi@foxmail.com);}
		\thanks{Tianrui~Li   is with the School of Computing and Artificial Intelligence, Southwest Jiaotong University, Chengdu 611756, China (e-mail: trli@swjtu.edu.cn).}
	}
	
	\markboth{Journal of \LaTeX\ Class Files,~Vol.~14, No.~8, August~2021}%
	{Shell \MakeLowercase{\textit{et al.}}: Feature and Instance Co-selection with  Multi-view Unsupervised Data  Reconstruction Learning}
	
	\IEEEpubid{0000--0000/00\$00.00~\copyright~2021 IEEE}
	
	\maketitle
	
	\begin{abstract}
	The objective of multi-view unsupervised feature and instance co-selection is to simultaneously identify the most representative features and samples from multi-view unlabeled data, which aids in mitigating the curse of dimensionality and reducing instance size to improve the performance of downstream tasks. However, existing methods treat feature selection and instance selection as two separate processes, failing to leverage the potential interactions between the feature and instance spaces. Additionally, previous co-selection methods for multi-view data require concatenating different views, which overlooks the consistent information among them. In this paper, we propose a CONsistency and DivErsity learNing-based multi-view unsupervised Feature and Instance co-selection (CONDEN-FI) to address the above-mentioned issues. Specifically, CONDEN-FI reconstructs multi-view data from both the sample and feature spaces to learn representations that are consistent across views and specific to each view, enabling the simultaneous selection of the most important features and instances. Moreover, CONDEN-FI adaptively learns a view-consensus similarity graph to help select both dissimilar and  similar samples in the reconstructed data space, leading to a more diverse selection of instances. An efficient algorithm is developed to solve the resultant optimization problem, and the comprehensive experimental results on real-world datasets demonstrate that CONDEN-FI is effective compared to state-of-the-art methods.
	\end{abstract}

	
	\section{Introduction} \label{Introduction}
	\IEEEPARstart{M}{ulti-view} data, which describe the same object with heterogeneous features derived from different kinds of sources, are prevalent in many real-world applications. For example, in medical diagnosis, each patient is evaluated from multiple perspectives, including imaging features from MRI scans, physiological readings from EEG measurements, and genetic traits from genomic analyses~\cite{MSAE}. In image classification, each image is described by various feature descriptors from different aspects, including Histogram of Oriented Gradients (HOG), Local Binary Patterns (LBP), Color Histograms (CH), and Scale-Invariant Feature Transform (SIFT)~\cite{MVMIML}. In those applications, heterogeneous features derived from various views in multi-view data are typically high-dimensional, resulting in the curse of dimensionality problem. Additionally, labeling a large number of data samples is time-consuming and even infeasible in practice. 	Multi-view unsupervised feature selection (MUFS) tackles this problem by selecting the most informative features from unlabeled data, thereby reducing the number of features and enhancing predictive performance~\cite{MUFSsurvey, EMUFS, SCMvFS}. Although MUFS has been extensively developed, it is often affected by redundant or noisy samples, which makes it challenging to identify important features and leads to a decrease in predictive performance~\cite{UFI,sCOs2}. Hence, instance selection, as the dual problem of feature selection, is crucial for enhancing predictive performance by selecting representative samples and reducing the number of samples typically considered noise or outliers~\cite{ISsurvey1, BSFRS}. However, instance selection is also influenced by irrelevant features~\cite{ISsurvey2, BSNID}. For example, in instance selection for medical diagnosis, irrelevant features like eye color can obscure the influence of important features such as blood pressure and glucose levels. This can lead to the selection of samples that do not accurately represent critical health indicators, thereby reducing the model's predictive performance. Considering that feature selection and instance selection affect each other,  a key question is how to simultaneously select representative features and samples by leveraging the interaction of feature and instance spaces to improve performance.

	In the literature, there are two distinct ways for jointly selecting features and instances. The first class of approaches uses a combination of feature selection and instance selection methods to choose features and instances sequentially from the data. This kind of method treats feature selection and instance selection as two separate processes, failing to utilize the potential interactions between the feature and instance spaces, which  results in the selected data being unrepresentative and significantly hinders the performance of downstream tasks. Rather than selecting features and instances sequentially, the second class of methods simultaneously selects both directly from the data. Typical methods in this class include Unified criterion for Feature and Instance selection (UFI)~\cite{UFI}. UFI jointly selects informative features and instances by minimizing the covariance matrix on the selected data, thereby reducing the prediction error. Besides, Du et al. proposed an unsupervised dual learning framework for the co-selection of features and instances, which reconstructs the latent data space using the selected features and instances while simultaneously preserving the global structure of the data~\cite{DFIS}. Furthermore, Benabdeslem et al. introduced a co-selection method that preserves data similarity by applying $\ell_{2,1-2}$-norm regularization. This approach enhances sparsity and eliminates unimportant features and samples when selecting a representative data subset~\cite{sCOs2}.  However, previous methods for co-selecting features and instances are limited to single-view data. When applied to multi-view data, different views must be concatenated before selecting features and instances, which overlooks the potential consistency information among the views. Additionally, these methods fail to learn a similarity graph that guides the selection of dissimilar and similar samples, which is beneficial for achieving diversity in the chosen representative instances.
	
\begin{figure*}[!htbp]
		\centering 
		\includegraphics[width=1\textwidth]{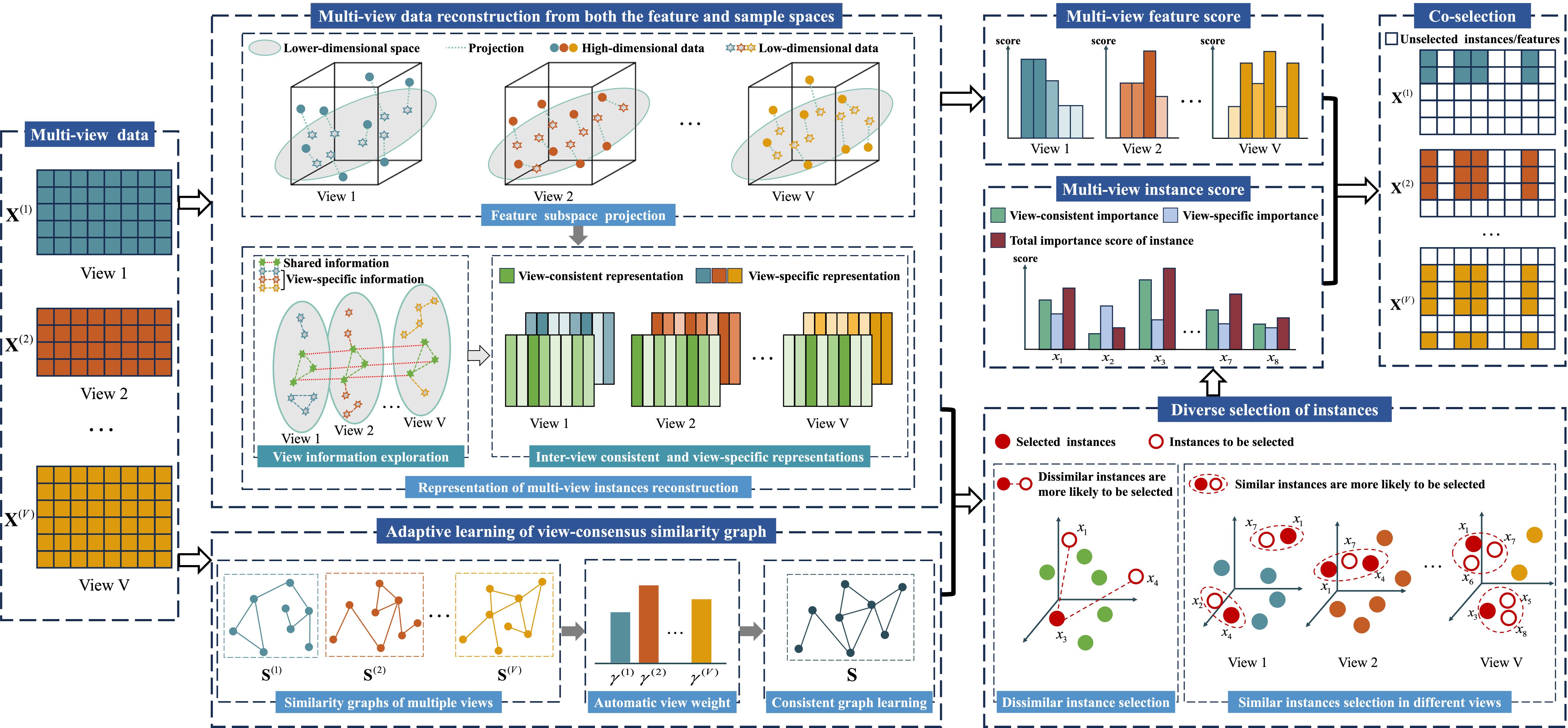}
		\caption{The framework of the proposed consistency and diversity learning-based multi-view unsupervised feature and instance co-selection (CONDEN-FI).}
		\label{overall framework}
	\end{figure*}

	To address the above  issues, we present a novel joint feature and instance selection method for multi-view unlabeled data, named  CONsistency and DivErsity learNing-based multi-view unsupervised Feature and Instance co-selection (CONDEN-FI).  Different from conventional methods that concatenate multi-view data and then perform feature and instance selection separately, we propose learning the inter-view consistent and view-specific representations within a reduced-dimensional space to reconstruct the original high-dimensional data, thereby enabling the simultaneous selection of  features and samples. Meanwhile, we adaptively learn a view-consensus similarity graph to more effectively select both dissimilar and similar samples in the reconstructed low-dimensional data space, which leads to a more diverse selection of instances. Moreover, a novel instance selection scoring measure is developed based on the importance of samples in both the inter-view consistent and view-specific representations of reconstructed multi-view data. We develop an alternative optimization algorithm to solve the proposed model and provide an analysis of its time complexity and convergence. Fig.~\ref{overall framework} illustrates the framework of the proposed CONDEN-FI. The major contributions of this paper are summarized as follows:
	
	\begin{itemize}
		\item To the best of our knowledge, this is the first work to address the problem of unsupervised feature and instance co-selection on multi-view unlabeled data by simultaneously utilizing both shared and view-specific information across different views.  
		\item We propose a novel feature and instance joint selection model by leveraging both view-consistent and view-specific representations to reconstruct the reduced-dimensional data space. Additionally, an adaptive view-consensus similarity graph learning module is integrated into this model to facilitate the diverse selection of instances.
		\item An effective iterative optimization algorithm with fast convergence is developed to solve the proposed model. Extensive experimental results demonstrate that the proposed method outperforms several state-of-the-art single-view and combined multi-view co-selection methods.
	\end{itemize}
	
	The rest of this paper is organized as follows. In Section 2, we briefly review related work on MUFS, instance selection, and feature and instance co-selection. In Section 3, we formulate CONDEN-FI and present an effective algorithm to solve this model, along with analyzing its time complexity and convergence. A series of experiments are carried out, and the comparative analysis of the results is detailed in Section 4. Section 5 shows the conclusions. 
	
	\section{Related Work} \label{Related Work}
	In this section, we review some representative works on multi-view unsupervised feature selection, instance selection, and the joint selection of both features and instances.
	
	\subsection{Multi-view Unsupervised Feature Selection}
	MUFS aims to identify a compact subset of representative features from multi-view data by leveraging latent information across different views. It has been widely studied and can be classified into two main categories. Methods in the first category begin by combining multi-view data into a single view, followed by applying unsupervised feature selection approaches designed for single-view data, such as Lapscore~\cite{Lapscore}, LRDOR~\cite{LRDOR} and FSDSC~\cite{FSDSC}. Lapscore selects the most informative features based on the Laplacian Score, which evaluates the ability of features to preserve locality. LRDOR integrates data structure learning and feature orthogonalization for achieving a low-redundancy in unsupervised feature selection. However, these single-view-based methods merely concatenate multi-view data, failing to utilize the latent correlations  across different views, resulting in inferior feature selection performance. To address this issue, the other category of methods selects features directly from multi-view data by utilizing the hidden information across different views. Zhang et al. explored view-within and view-across information through view-specific and joint projections and integrated these into an adaptive graph learning-based multi-level projection framework to select representative features from multi-view data~\cite{MAMFS}. MFSGL~\cite{MFSGL} adopts an adaptive weighting method to measure the importance of each view and learns an optimal consensus similarity matrix across different views during feature selection. CDMvFS~\cite{CDMvFS} learns multiple similarity graphs equipped with view diversity regularization from different views and incorporates them into a consensus clustering framework to facilitate discriminative feature selection. Although MUFS has been extensively investigated, it is often affected by redundant or noisy samples, making it difficult to identify important features. Selecting representative instances from multi-view data during preprocessing is important to collaboratively enhancing feature selection and boosting the performance of downstream tasks.
	
	\subsection{Instance Selection}
	Instance selection aims to select the most informative samples from the original data and eliminate instances typically regarded as noise or outliers, which makes the selected data more suitable for model training and enhances the model's performance. Based on label availability, instance selection can be classified into supervised and unsupervised categories. As a typical supervised method, EGDIS employs a global density function to quantify the number of neighbors and a correlation function to assess the consistency of these neighbors when selecting representative instances~\cite{EGDIS}. Gong et al. incorporated neighbor label information into the evidence theory framework to select boundary instances and filter out noise~\cite{EIS}. Liu et al. proposed a memory-based probabilistic ranking method to identify noisy labels and select informative instances by utilizing features extracted from neural network models~\cite{PRISM}. Since acquiring a large number of data labels is costly and time-consuming, unsupervised instance selection methods have been developed to identify the most informative instances from unlabeled data. Dornaika et al. proposed a decremental sparse modeling approach for unsupervised instance selection, which utilizes staged representative prototypes and pruning to mitigate the impact of noisy samples~\cite{D-SMRS}. CIS uses K-Means clustering to identify clusters in the unlabeled data and then selects instances from the centers and borders of these clusters using a distance-based measure~\cite{CIS}. Aydin employed the data scaling technique to divide the data into multiple conjectural hyper-rectangles, keeping one instance in each hyper-rectangle and discarding the others to select representative instances~\cite{NIS}. However, these unsupervised methods can be impacted by irrelevant or redundant features, which makes it more difficult to select accurately representative samples. Conversely, feature selection can also be affected by instance selection. Therefore, there is an urgent need to develop a joint feature and instance selection method to improve the  performance of downstream tasks.
	
	\subsection{Joint Selection of Features and Instances}
	The joint selection of features and instances can be classified into two categories. The first category combines feature selection and instance selection to select both features and instances sequentially. Approaches in this category treat feature selection and instance selection as separate processes, neglecting the interactions between the feature space and the instance space. To address this issue, the second category simultaneously selects both features and instances directly from the entire dataset. Tan et al. utilized the link information found in social media data and incorporated it into the instance selection process along with feature selection~\cite{coselect}. Zhang et al. used fuzzy rough sets to determine the importance degree of fuzzy granules and selected representative features and instances based on how well they preserved that importance degree~\cite{BSFRS}. Kusy et al. combined random forest, clustering, and nearest neighbor techniques to achieve the joint selection of features and instances~\cite{FSIS}. BSNID introduces a conditional neighborhood entropy measure designed to quantify the decision uncertainty of neighborhood granules, which facilitates the selection of representative features and samples~\cite{BSNID}. The aforementioned methods depend on labeled data, but obtaining a substantial number of labels can be time-consuming and infeasible in real-world applications. Zhang et al. proposed an unsupervised joint selection of features and instances by minimizing the covariance matrix to reduce prediction error~\cite{UFI}. DFIS reconstructs the latent space of data from a reduced subset, which preserves global structure for dual selection of features and instances~\cite{DFIS}. Benabdeslem et al. proposed a similarity-preserving method equipped with sparse regularization using the $\ell_{2,1-2}$ norm to simultaneously select features and instances in unsupervised and semi-supervised scenarios~\cite{sCOs2}. However, these methods cannot be directly applied to multi-view data, as they require combining the data into a single view for feature and instance selection, which overlooks the potential information present between views. To the best of our knowledge, there is still no work that utilizes the shared and view-specific information across views for the co-selection of features and instances in multi-view unlabeled data.

	\section{Proposed Method}
	
	\subsection{Notations} \label{Notations}
	In this paper, matrices and vectors are written in boldface uppercase. For a given matrix $\mathbf{X} \in \mathbb{R}^{d \times n}$, its $i$-th row and $j$-th column are denoted as $\mathbf{X}_{i.}$ and $\mathbf{X}_{.j}$, respectively. The ($i$, $j$)-th entry of  $\mathbf{X}$ is represented as $x_{ij}$. Let $\operatorname{Tr}(\mathbf{X})$ denote the trace of $\mathbf{X}$, and $\mathbf{X}^T$ represent its transpose. $\mathbf{1}_{n}$ is a column vector of size $n\times 1$ with all entries equal to one (denoted by $\mathbf{1}$ if the size is clear from the context). The Frobenius norm of  matrix $\mathbf{X}$ is defined as $\| \mathbf{X}\|_F=\sqrt{\sum_{i=1}^{d}\sum_{j=1}^{n}x_{i j}^{2}}$. The $\ell_{2,1}$-norm of $\mathbf{X}$, denoted as $\|\mathbf{X}\|_{2,1}$, is given by $\sum_{i=1}^{d} \sqrt{\sum_{j=1}^{n} \mathbf{X}_{i j}^{2}}=\sum_{i=1}^{d}\| \mathbf{X}_{i\cdot} \|_{2}$, where $\|\mathbf{X}_{i\cdot}\|_{2}$ is  the $\ell_{2}$-norm of vector $\mathbf{X}_{i\cdot}$.
	
	Let $\mathcal{X}=\{\mathbf{X}^{(v)},v=1,\ldots,V\}$ represent  multi-view data with $V$ views,  where $\mathbf{X}^{(v)} \in \mathbb{R}^{d_{v} \times n}$ is the data matrix of the $v$-th view, containing $d_v$ features and $n$ instances. Our objective  is to simultaneously select $l$ discriminative features and $m$ informative instances from the multi-view data $\mathcal{X}$.
	
	\subsection{Formulation of CONDEN-FI} \label{Methodology}
	Data self-representation, where each instance is expressed as a linear combination of others, has been widely used in fields like clustering and anomaly detection~\cite{SSC,LSR, R-graph}. 
	In this paper, data self-representation is used to select representative instances from multi-view data by reconstructing the original data in each view. This can be formulated as follows:
	\begin{equation} \label{New1}
		\begin{aligned}
			\min_{\tiny{\mathbf{B}^{(v)}}}&\sum_{v=1}^{V}{(\|\mathbf{X}^{(v)}-\mathbf{X}^{(v)}\mathbf{B}^{(v)}\|_F^{2}+{\eta^{(v)}}{{\| {\mathbf{B}^{(v)}}\|}_{2,1}})},\\
		\end{aligned}
	\end{equation}
	where $\mathbf{B}^{(v)} \in R^{n \times n}$ is an instance selection matrix and $\eta^{(v)}$ is a trade-off parameter. The $i$-th row of $\mathbf{B}^{(v)}$ represents the contribution of the $i$-th instance to the reconstruction of all other instances, thus indicating the importance of the $i$-th instance in $\mathbf{X}^{(v)}$. A  $\ell_{2,1}$-norm is imposed on the  instance selection matrix $\mathbf{B}^{(v)}$ to encourage the weights of unimportant instances approach zero. Then, the representative instances can be obtained through computing the $\ell_{2}$-norm of each row of $\mathbf{B}^{(v)}$ and selecting the top $m$ ranked instances in descending order. Eq.~(\ref{New1}) approximates the original data by selecting important instances in each view, aiming to minimize the reconstruction error across the entire multi-view dataset. However, Eq.~(\ref{New1}) treats each view of the multi-view data independently, ignoring the common information among the views. To capture the shared information for selecting a representative subset from multi-view data, we divide the instance selection matrix into a consistent part and a view-specific part, which  can be expressed as follows:
	\begin{equation} \label{New2}
		\begin{aligned}
			\min_{\tiny{\mathbf{B}^{(v)}, \mathbf{B}, \eta^{(v)}}}&\sum_{v=1}^{V}{(\|\mathbf{X}^{(v)}\!-\!\mathbf{X}^{(v)}(\mathbf{B}+\mathbf{B}^{(v)})\|_F^{2}\!+\!{\eta^{(v)}}^{r}{{\| {\mathbf{B}^{(v)}}\|}_{2,1}})}\\
			&+\theta {{\| \mathbf{B}\|}_{2,1}}\\
			s.t.~ & \sum_{v=1}^{V}{\eta}^{(v)}=1,\quad {\eta}^{(v)}\geq 0,
		\end{aligned}
	\end{equation}
	where $\mathbf{B}$ and $\mathbf{B}^{(v)}$ respectively denote the inter-view consistent and view-specific instance selection matrices, $r$ is a trade-off parameter, and $\theta$ is a regularization parameter that controls the sparsity of $\mathbf{B}$. In Eq.~(\ref{New2}), the original multi-view data $\mathbf{X}^{(v)}$ is approximated by a view-consistent representation $\mathbf{X}^{(v)}\mathbf{B}$ that captures shared information across different views and a view-specific representation $\mathbf{X}^{(v)}\mathbf{B}^{(v)}$ that captures unique information for each view. We can then select representative instances using $\mathbf{B}$ and $\mathbf{B}^{(v)}$ to best approximate the original multi-view data. To this end, a novel instance selection scoring measure is proposed as follows:
	\begin{equation} \label{New3}
		MvIS_{i} = \|\mathbf{B}_{i\cdot}\|_{2}^{2} + \sum_{v=1}^{V}{\eta^{(v)}\frac{1}{\|\mathbf{B}^{(v)}_{i\cdot}\|_{2}^{2}}}~(i=1,2,...,n),
	\end{equation}
	where $MvIS_i$ represents the importance score of the $i$-th instance. We use Eq.~(\ref{New3}) to calculate the score for each instance, and then select the top $m$ instances with the highest scores. In Eq.~(\ref{New3}), $\mathbf{B}_{i\cdot}$ and $\mathbf{B}^{(v)}_{i\cdot}$ represent the importance of the $i$-th instance in the common representation and the $v$-th view-specific representation, respectively. Eq.~(\ref{New3}) implies that if the $i$-th sample plays a more significant role in reconstructing the common representation of multi-view data, it is more likely to be selected. Conversely, if the $i$-th sample is more important for reconstructing the view-specific representation, it is less likely to be chosen. Through Eq.~(\ref{New3}), we can select samples that embody the common representation across different views, rather than those that reflect only the view-specific representation. Besides, using Eq.~(\ref{New2}), we can adaptively determine the weight $\eta^{(v)}$ for each view-specific instance selection matrix without requiring manual settings for each view.
	
	Furthermore, considering that redundant and noisy features can affect the reconstruction error of selected instances approximating the original data, we use a transformation matrix $\mathbf{W}^{(v)} \in \mathbb{R}^{d_v \times c}$ in each view to project the original high-dimensional feature space into a lower-dimensional space to select informative features. This  can be formulated as follows:
	\begin{equation} \label{New4}
		\begin{aligned}
			\min_{\tiny{\substack{\mathbf{B}^{(v)},\mathbf{B},\eta^{(v)},\\\mathbf{W}^{(v)},\lambda^{(v)}}}}&\sum_{v=1}^{V}{(\| {{\mathbf{W}^{(v)}}^T }{\mathbf{X}^{(v)}}-{{\mathbf{W}^{(v)}}^T }{\mathbf{X}^{(v)}}( \mathbf{B}+{\mathbf{B}^{(v)}} ) \|_{F}^{2}  }\\
			&+{{\eta^{(v)}}^{r}{{\| {\mathbf{B}^{(v)}} \|}_{2,1}}+\lambda^{(v)}}^{r}{{\| {\mathbf{W}^{(v)}} \|}_{2,1}})
			+\theta {{\| \mathbf{B} \|}_{2,1}} \\
			s.t.~&\!({\mathbf{W}^{(v)}}^T\mathbf{X}^{(v)})\!({\mathbf{W}^{(v)}}^T\mathbf{X}^{(v)})\!^{T}\!\!\!=\!\!\mathbf{I}, \sum_{v=1}^{V}\!\!{\lambda}^{(v)}\!\!=\!\!1,{\lambda}^{(v)}\!\!\geq\!\! 0,\\
			&\sum_{v=1}^{V}{\eta}^{(v)}\!=\!1,{\eta}^{(v)}\!\geq\!0,
		\end{aligned}
	\end{equation}
	where ${\lambda}^{(v)}$ is a regularization parameter. The orthogonal constraint in Eq.~(\ref{New4}) ensures that the projected data representations are uncorrelated while avoiding the trivial solution. By calculating the $\ell_{2}$-norm of each row of $\mathbf{W}^{(v)}$ in every view and sorting them in descending order, we can then select the top $l$ most informative features. Eq.~(\ref{New4}) enables the simultaneous selection of features and samples, effectively reducing feature dimensionality and selecting representative instances from inter-view and view-specific aspects, thereby improving the quality of the selected subset.

	Previous works have demonstrated that exploring and utilizing the local manifold structure of data is beneficial to improve the performance of multi-view unsupervised learning ~\cite{GLSPFS, CVLPDCL}. In this paper, we also maintain the local manifold structure of multi-view data by learning a consensus similarity graph  from different views, which is formulated as follows:
	\begin{equation} \label{New5}
		\begin{aligned}
			\min_{\tiny{\substack{ \mathbf{S}, \mathbf{B}^{(v)}, \\\gamma^{(v)}}}}
			& \sum_{v=1}^{V}\!{\gamma^{(v)}}^{r}{\| \mathbf{S}\!-\!\mathbf{S}^{(v)}\|_{F}^2}\!+\!\frac{\alpha}{2}\sum_{v=1}^{V}\sum_{i,j=1}^{n}
			\| \mathbf{B}^{(v)}_{i\cdot}\!-\!\mathbf{B}^{(v)}_{j\cdot}\|_{2}^{2} s_{ij}^{(v)}  \\
			s.t.~& \mathbf{S}_{i\cdot}\mathbf{1}=1, s_{ij} \geq 0, \sum_{v=1}^{V}{\gamma}^{(v)}=1, {\gamma}^{(v)}\geq 0,
		\end{aligned}
	\end{equation}
	where $\mathbf{S}$ is the consensus similarity-induced graph matrix to be learned automatically, $\mathbf{S}^{(v)}$ is the similarity matrix of the $v$-th view constructed by $k$-nearest neighbors~\cite{GMC}, $\gamma^{(v)}$ is an adaptive view-weight, and $\alpha$ is a trade-off hyper-parameter. The first term in Eq.~(\ref{New5})  ensures that the similarity matrices of different views are close to a common consensus similarity matrix. The second term guarantees  that the more similar two samples are, the closer their corresponding contributions will be when constructing the view-specific representation. Through Eq.~(\ref{New5}), we can leverage the consistent manifold structure across different views to enhance the performance of  feature and instance co-selection.
	
	In addition, by utilizing the learned common similarity matrix, we regularize that less similar samples contribute more equally  to the reconstruction of view-consistent representation. This increases the possibility of simultaneously selecting dissimilar samples, which achieves a diverse selection of instances. The regularization  can be formulated  as follows:
	\begin{equation} \label{New6}
		\begin{aligned}
			\underset{\mathbf{B},\mathbf{S}}{\mathop{\min }}
			& \sum_{i,j=1}^{n}
			\| \mathbf{B}_{i\cdot} - \mathbf{B}_{j\cdot} \|_{2}^{2} (1-s_{ij})  \\
			s.t.~& \mathbf{S}_{i\cdot}\mathbf{1}=1, s_{ij} \geq 0.
		\end{aligned}
	\end{equation}
	
	By combining Eqs.~(\ref{New4}), (\ref{New5}), and (\ref{New6}) together, the proposed consistency and diversity learning-based multi-view unsupervised feature and instance co-selection method (CONDEN-FI) is summarized as follows:
	\begin{equation} \label{NewObjFunc}
		\begin{aligned}
			\min _{\Omega} &\sum_{v=1}^{V}{(\| {{\mathbf{W}^{(v)}}^T }{\mathbf{X}^{(v)}}-{{\mathbf{W}^{(v)}}^T }{\mathbf{X}^{(v)}}( \mathbf{B}+{\mathbf{B}^{(v)}} ) \|_{F}^{2}  }\\
			&+{{\eta^{(v)}}^{r}{{\| {\mathbf{B}^{(v)}} \|}_{2,1}}+\lambda^{(v)}}^{r}{{\| {\mathbf{W}^{(v)}} \|}_{2,1}})
			+\theta {{\| \mathbf{B} \|}_{2,1}}\\
			&+\frac{\alpha}{2}\!\sum_{i,j=1}^{n}{\![
				\| \mathbf{B}_{i\cdot}\!-\!\mathbf{B}_{j\cdot} \|_{2}^{2} (1\!-\!s_{ij})
				\!+\!\!\sum_{v=1}^{V}{\!\| \mathbf{B}^{(v)}_{i\cdot}\!-\!\mathbf{B}^{(v)}_{j\cdot} \|_{2}^{2} s_{ij}^{(v)}}
				]} \\
			&+\sum_{v=1}^{V}\!{\gamma^{(v)}}^{r}{\| \mathbf{S}\!-\!\mathbf{S}^{(v)}\|_{F}^2}\! \\
			s.t.&~({\mathbf{W}^{(v)}}^T\mathbf{X}^{(v)})({\mathbf{W}^{(v)}}^T\mathbf{X}^{(v)})^{T}=\mathbf{I}, \mathbf{S}_{i\cdot}\mathbf{1}=1, s_{ij} \geq 0,\\
			&~\boldsymbol{\Psi}^{(v)}\geq 0, \sum_{v=1}^{V}\boldsymbol{\Psi}^{(v)}=\mathbf{1}_{3}, \\
		\end{aligned}
	\end{equation}
	where $\Omega=\{\mathbf{W}^{(v)}, \mathbf{B}, \mathbf{B}^{(v)}, \mathbf{S}, \boldsymbol{\Psi}^{(v)}\}$ and $\boldsymbol{\Psi}^{(v)}=[\lambda^{(v)}, \eta^{(v)}, \gamma^{(v)}]^T$.
	
	Eq.~(\ref{NewObjFunc}) shows that the proposed CONDEN-FI offers two advantages. First, CONDEN-FI uses a self-representation technique to split the multi-view sample space into two parts: one that reconstructs the shared representation across different views, and another that captures the unique representation of each individual view. By selecting samples that significantly aid in reconstructing consistent information and choosing features with high discriminative power in the projection space, a more representative subset of multi-view data can be effectively achieved. Second, by adaptively learning an inter-view consistency similarity matrix with a regularization term for sample diversity selection, CONDEN-FI can select more diverse samples and better preserve the local manifold structure of multi-view data compared to using a predefined similarity matrix.
	
	\subsection{Solution of CONDEN-FI}
	It can be seen that the variables in the objective function of Eq.~(\ref{NewObjFunc}) are interrelated, making it difficult to solve them simultaneously. Hence, we propose an alternative iterative algorithm to address the optimization problem by optimizing the objective function for one variable at a time, while keeping the other variables fixed. 
	
	\subsubsection{Updating $\mathbf{W}^{(v)}$ with Other Variables Fixed}
	When the other variables are fixed and irrelevant terms are removed, the problem of solving for $\mathbf{W}^{(v)}$ reduces to the following formulation:
	\begin{equation} \label{NewUpdateWv}
		\begin{aligned}
			\underset{\mathbf{W}^{(v)}}{\mathop{\min }}
			&\| {{\mathbf{W}^{(v)}}^T }{\mathbf{X}^{(v)}}\!\!-\!\!{{\mathbf{W}^{(v)}}^T }{\mathbf{X}^{(v)}}\!( \mathbf{B}\!+\!\!{\mathbf{B}^{(v)}}\!)\|_{F}^{2}\!+\!\!{\lambda^{(v)}}^{r}{{\| {\mathbf{W}^{(v)}}\|}_{2,1}}\\
			s.t.~&{\mathbf{W}^{(v)}}^{T}\mathbf{X}^{(v)} { {\mathbf{X}^{(v)}}^{T}\mathbf{W}^{(v)}}=\mathbf{I}.
		\end{aligned}
	\end{equation}
	
	Based on the properties of  matrix trace and $\ell_{2,1}$-norm~\cite{RFS}, Eq.~(\ref{NewUpdateWv}) can be transformed into the following form:
	\begin{equation} \label{NewWvProblem}
		\begin{aligned}
			\underset{{\mathbf{W}^{(v)}}}{\mathop{\min }} 
			&\operatorname{Tr}({\mathbf{W}^{(v)}}^{T}{\mathbf{X}^{(v)}}{\mathbf{H}^{(v)}}{\mathbf{X}^{(v)}}^{T}\mathbf{W}^{(v)})\\
			&+{\lambda ^{(v)}}^{r} \operatorname{Tr}( {\mathbf{W}^{(v)}}^{T }{\mathbf{D}_{v_{1}}}{\mathbf{W}^{(v)}} ) \\
			s.t.~&{\mathbf{W}^{(v)}}^{T}\mathbf{X}^{(v)} { {\mathbf{X}^{(v)}}^{T}\mathbf{W}^{(v)}}=\mathbf{I},
		\end{aligned}
	\end{equation}
	where $\mathbf{H}^{(v)} =(\mathbf{I}-\mathbf{B}-{\mathbf{B}^{(v)}} ) {(\mathbf{I}- \mathbf{B}-{\mathbf{B}^{(v)}})^T}$, and $\mathbf{D}_{v_{1}}$ is a diagonal matrix with its $i$-th diagonal element given by $\frac{1}{2{{\| {\mathbf{W}^{(v)}_{i\cdot}} \|}_{2}}+\varepsilon}$ ( $\varepsilon$ is a relatively small value used to avoid a zero denominator).
	
	Following \cite{FSASL, updateW2}, the optimal solution for $\mathbf{W}^{(v)}$ can be obtained as follows:
	\begin{equation} \label{NewupdateWv1}
		\mathbf{W}^{(v)} = (\mathbf{X}^{(v)}{\mathbf{X}^{(v)}}^{T} + {\lambda^{(v)}}^{r}\mathbf{D}_{v_{1}})^{-1}\mathbf{X}^{(v)}\mathbf{Y}^{(v)}, 
	\end{equation}
	where $\mathbf{Y}^{(v)}$ is the eigenvector matrix of $\mathbf{H}^{(v)}$.  In the scenario where  $d_v > n$, we use the method from \cite{updateW1} to reduce the computational complexity associated with the matrix inverse calculation in Eq.~(\ref{NewupdateWv1}). By leveraging the  Woodbury matrix identity, Eq.~(\ref{NewupdateWv1}) can be transformed into: 
	\begin{equation} \label{NewUpdateWv2}
		\mathbf{W}^{(v)} = {\lambda^{(v)}}^{-r} \mathbf{R}^{(v)} \mathbf{X}^{(v)}\mathbf{Y}^{(v)},
	\end{equation} 
	where 
	$\mathbf{R}^{(v)} = \mathbf{D}_{v_{1}}^{-1} -{\lambda^{(v)}}^{-r}\mathbf{D}_{v_{1}}^{-1}\mathbf{X}^{(v)}\mathbf{O}^{(v)}{\mathbf{X}^{(v)}}^T\mathbf{D}_{v_{1}}^{-1}$ and 
	$\mathbf{O}^{(v)}  = (\mathbf{I}_{n}+{\lambda^{(v)}}^{-r}{\mathbf{X}^{(v)}}^T\mathbf{D}_{v_{1}}^{-1}\mathbf{X}^{(v)})^{-1}$.
	
	\subsubsection{Updating $\mathbf{B}$ with Other Variables Fixed}
	When other variables are fixed, we can update $\mathbf{B}$ by solving the following problem:
	\begin{equation} \label{NewUpdateB}
		\begin{aligned}
			\underset{{\mathbf{B}}}{\mathop{\min }}
			&\sum\limits_{v=1}^{V}{\| {\mathbf{W}^{(v)}}^{T}{\mathbf{X}^{(v)}}-{\mathbf{W}^{(v)}}^{T}{\mathbf{X}^{(v)}}( \mathbf{B}+{\mathbf{B}^{(v)}} ) \|_{F}^{2}} \\
			&+\frac{\alpha}{2}\sum_{i,j=1}^{n}{\| \mathbf{B}_{i\cdot}-\mathbf{B}_{j\cdot} \|_{2}^{2} (1-s_{ij})}+\theta {{\| \mathbf{B} \|}_{2,1}}.
		\end{aligned}
	\end{equation}
	
	Given that $\operatorname{Tr}({\mathbf{B}^T}{\mathbf{L}_{{\bar{S}}}}\mathbf{B})\!\!=\!\!\frac{\alpha}{2}\sum_{i,j=1}^{n}{\| \mathbf{B}_{i\cdot}\!\!-\!\!\mathbf{B}_{j\cdot} \|_{2}^{2} (1\!\!-\!\!s_{ij})}$, where $\mathbf{L}_{{\bar{S}}}$ is the Laplacian matrix of $\bar{S}$ with the $(i,j)$-th entry being $1\!\!-\!\!s_{ij}$, Eq.~(\ref{NewUpdateB}) can be rewritten as follows:
	\begin{equation} \label{NewUpdateB2}
		\begin{aligned}
			\underset{{\mathbf{B}}}{\mathop{\min }}
			&\sum\limits_{v=1}^{V}{\| {\mathbf{W}^{(v)}}^{T}{\mathbf{X}^{(v)}}-{\mathbf{W}^{(v)}}^{T}{\mathbf{X}^{(v)}}( \mathbf{B}+{\mathbf{B}^{(v)}} ) \|_{F}^{2}} \\
			&+\alpha \operatorname{Tr} ( {\mathbf{B}^T}{\mathbf{L}_{{\bar{S}}}}\mathbf{B})+\theta \operatorname{Tr} ( {\mathbf{B}^T}{\mathbf{D}_B\mathbf{B}}),
		\end{aligned}
	\end{equation}
	where $\mathbf{D}_B$ is a diagonal matrix with  the $i$-th diagonal element defined as $\frac{1}{2{{\| \mathbf{B}_{i\cdot}\|}_{2}}+\varepsilon}$.
	
	By taking the partial derivative of the objective function in Eq.~(\ref{NewUpdateB2}) with regards to $\mathbf{B}$ and setting it to zero, we have:
	\begin{equation} \label{NewEqB}
		\begin{aligned}
			&\sum\limits_{v=1}^{V}{\mathbf{X}^{(v)}}^T{\mathbf{W}^{(v)}}{\mathbf{W}^{(v)}}^T{\mathbf{X}^{(v)}}{\left( \mathbf{B}+{\mathbf{B}^{(v)}}-\mathbf{I} \right)} \\
			&+\alpha {\mathbf{L}_{{\bar{S}}}}\mathbf{B}+\theta {\mathbf{D}_{B}}\mathbf{B}=\mathbf{0}.
		\end{aligned}
	\end{equation}
	
	Then, according to Eq.~(\ref{NewEqB}), we get 
	\begin{equation} \label{NewupdateB}
		\mathbf{B} = 
		\mathbf{G}_B
		[\sum\limits_{v=1}^{V}{\mathbf{X}^{(v)}}^T{\mathbf{W}^{(v)}}{\mathbf{W}^{(v)}}^{T}\mathbf{X}^{(v)}(\mathbf{I}-\mathbf{B}^{(v)})],
	\end{equation}
	where $\mathbf{G}_{B} = (\sum\limits_{v=1}^{V}{{\mathbf{X}^{(v)}}^{T}{\mathbf{W}^{(v)}}{\mathbf{W}^{(v)}}^{T}\mathbf{X}^{(v)}} + \alpha {\mathbf{L}_{{\bar{S}}}}+\theta {\mathbf{D}_{B}})^{-1}$.

	\subsubsection{Updating $\mathbf{B}^{(v)}$ with Other Variables Fixed}
	When fixing other variables and removing the irrelevant terms, solving $\mathbf{B}^{(v)}$ reduces to the following problem:
	\begin{equation} \label{NewUpdateBv}
		\begin{aligned}
			\underset{{\mathbf{B}^{(v)}}}{\mathop{\min }}
			&\| {\mathbf{W}^{(v)}}^{T}{\mathbf{X}^{(v)}}-{\mathbf{W}^{(v)}}^{T}{\mathbf{X}^{(v)}}( \mathbf{B}+{\mathbf{B}^{(v)}} ) 
			\|_{F}^{2} \\
			&+\frac{\alpha}{2} \sum_{i,j=1}^{n}{\|  \mathbf{B}^{(v)}_{i\cdot}-\mathbf{B}^{(v)}_{j\cdot}\|_{2}^2 s_{ij}^{(v)}}
			+{\eta^{(v)}}^{r}{{\| \mathbf{B}^{(v)} \|}_{2,1}}
		\end{aligned}
	\end{equation}
	
	Following the same process as solving Eq. (\ref{NewUpdateB}), we take the derivative of Eq. (\ref{NewUpdateBv}) with respect to $\mathbf{B}^{(v)}$ and set it to zero, allowing us to obtain:
	\begin{equation}\label{NewUpdateBv2}
		\begin{aligned}
			&({\mathbf{X}^{(v)}}^T\mathbf{W}^{(v)}{\mathbf{W}^{(v)}}^T\mathbf{X}^{(v)} +\alpha\mathbf{L}_{S^{(v)}} +{\eta^{(v)}}^{r}\mathbf{D}_{v_{2}}) \mathbf{B}^{(v)} \\
			&={\mathbf{X}^{(v)}}^T\mathbf{W}^{(v)}{\mathbf{W}^{(v)}}^T\mathbf{X}^{(v)} (\mathbf{I}-\mathbf{B}), 
		\end{aligned}
	\end{equation}
	where $\mathbf{L}_{S^{(v)}}$ is the Laplacian matrix of $S^{(v)}$, $\mathbf{D}_{v_{2}}$ is a diagonal matrix with its $i$-th diagonal entry given by $\frac{1}{2{{\| {\mathbf{B}^{(v)}_{i\cdot}} \|}_{2}}+\varepsilon }$. According to Eq.~(\ref{NewUpdateBv2}), we have 
	\begin{equation} \label{NewUpdateBv3}
		\mathbf{B}^{(v)} = \mathbf{G}_{B^{(v)}} [{\mathbf{X}^{(v)}}^T\mathbf{W}^{(v)}{\mathbf{W}^{(v)}}^T\mathbf{X}^{(v)} (\mathbf{I}-\mathbf{B})],
	\end{equation}
	where $\mathbf{G}_{B^{(v)}}=({\mathbf{X}^{(v)}}^T\mathbf{W}^{(v)}{\mathbf{W}^{(v)}}^T\mathbf{X}^{(v)}+\alpha\mathbf{L}_{S^{(v)}} +{\eta^{(v)}}^{r}\mathbf{D}_{v_{2}})^{-1}$.
	
	\subsubsection{Updating $\mathbf{S}$ with Other Variables Fixed}
	After fixing the other variables, the optimization problem for updating $\mathbf{S}$ becomes the following:
	\begin{equation} \label{NewUpdateS}
		\begin{aligned}
			\underset{\mathbf{S}}{\mathop{\min }}
			& \sum_{v=1}^{V}{{\gamma^{(v)}}^{r}{\|\mathbf{S}-\mathbf{S}^{(v)}\|_{F}^2}}-\frac{\alpha }{2}\sum_{i,j=1}^{n}{\| {\mathbf{B}_{i\cdot}}-{\mathbf{B}_{j\cdot}} \|_{2}^{2}{s}_{ij}}\\
			s.t.~&{{s}_{ij}}\ge 0,{\mathbf{S}_{i\cdot}}\mathbf{1}=1.
		\end{aligned}
	\end{equation}
	
	Since Eq.~(\ref{NewUpdateS}) is independent for different $i$, we can address the following equivalent problem for each $\mathbf{S}_{i\cdot}$ by completing the square for ${s}_{ij}$ and removing terms unrelated to ${s}_{ij}$.
	\begin{equation}   \label{NewUpdateS2}
		\begin{aligned}
			\underset{\mathbf{S}_{i\cdot}}{\mathop{\min }} 
			& \sum_{v=1}^{V}
			{\| \mathbf{S}_{i\cdot} - \mathbf{P}^{(v)}_{i\cdot}\|}_{2}^{2} \\
			s.t.~&{{s}_{ij}}\ge 0,{\mathbf{S}_{i\cdot}}\mathbf{1}=1,
		\end{aligned}
	\end{equation}
	where $\mathbf{P}^{(v)}_{i\cdot} = \mathbf{S}^{(v)}_{i\cdot} + \frac{\alpha}{4V{{\gamma}^{(v)}}^{r}}\mathbf{A}_{i\cdot}$, and the $j$-th element of $\mathbf{A}_{i\cdot}$ is defined as $a_{ij}=\| \mathbf{B}_{i\cdot}-\mathbf{B}_{j\cdot}\|_2^2$.
	
	Following \cite{GMC}, the optimal solution to Eq.~(\ref{NewUpdateS2}) can be obtained as follows:
	\begin{equation} \label{NewUpdateS3}
		s_{ij}=(q_{ij}-\hat{\phi}^{*})_{+},
	\end{equation}
	where $q_{ij}$ is the $j$-th element of $\mathbf{Q}_{i\cdot}=\frac{1}{V}\sum_{v=1}^{V}{\mathbf{P}^{(v)}_{i\cdot}} + \frac{1}{n}\mathbf{1}^{T} -\frac{1}{Vn}\sum_{v=1}^{V}{\mathbf{P}^{(v)}_{i\cdot}}\mathbf{1}\mathbf{1}^{T}$, and $\hat{\phi}^{*}$  is determined by applying the Newton method to iteratively find the root of the equation $\frac{1}{n}\sum_{j=1}^{n}{( \hat{\phi} -q_{ij})_{+}} -\hat{\phi}=0$. 
	
	\subsubsection{Updating $\boldsymbol{\Psi}^{(v)}$ with Other Variables Fixed}
	When other variables are fixed, Eq.~(\ref{NewObjFunc}) can be reformulated as:
	\begin{equation} \label{NewUpdateWeight}
		\begin{aligned}
			\underset{\boldsymbol{\Psi}^{(v)}}{\mathop{\min }} &~[{\| {\mathbf{W}^{(v)}} \|}_{2,1}, {\| {\mathbf{B}^{(v)}} \|}_{2,1}, \| \mathbf{S}\!-\!\mathbf{S}^{(v)}\|_{F}^2] (\boldsymbol{\Psi}^{(v)})^{\circ r}\\
			s.t. &~\boldsymbol{\Psi}^{(v)}\geq 0, \sum_{v=1}^{V}\boldsymbol{\Psi}^{(v)}=\mathbf{1}_{3}, \\
		\end{aligned}
	\end{equation}
	where $(\boldsymbol{\Psi}^{(v)})^{\circ r}$ denotes the element-wise exponentiation of $\boldsymbol{\Psi}^{(v)}$ to the power of $r$.
	
	If we initially disregard the non-negative constraints, the Lagrange function for Eq.~(\ref{NewUpdateWeight}) is given by
	\begin{equation}  \label{NewUpdateWeight2}
		[{\|\!{\mathbf{W}^{(v)}} \!\|}_{2,1}, {\|\!{\mathbf{B}^{(v)}}\!\|}_{2,1}, \|\! \mathbf{S}\!-\!\mathbf{S}^{(v)}\!\|_{F}^2]\!(\boldsymbol{\Psi}^{(v)})^{\circ r}
		\!-\!\boldsymbol{\psi}^{T}\!(\sum_{v=1}^{V}\boldsymbol{\Psi}^{(v)}\!-\!\mathbf{1}_{3}),
	\end{equation}
	where $\boldsymbol{\psi}$ is the Lagrange multiplier. By taking the partial derivatives of each element $\lambda^{(v)}$, $\eta^{(v)}$, and $\gamma^{(v)}$ in $\boldsymbol{\Psi}^{(v)}$, and setting them to zero, we obtain the solutions for these elements as follows:
	\begin{equation} \label{NewUpdateWeight3}
		\lambda^{(v)} = \frac{(\| \mathbf{W}^{(v)}\|_{2,1})^ {\frac{1}{1-r}}}{\sum_{v=1}^{V}{( \| \mathbf{W}^{(v)}\|_{2,1})^ {\frac{1}{1-r}}}}
	\end{equation}
	\begin{equation} \label{NewUpdateWeight4}
		\eta^{(v)} = \frac{( \| \mathbf{B}^{(v)}\|_{2,1})^ {\frac{1}{1-r}}}{\sum_{v=1}^{V}{( \| \mathbf{B}^{(v)}\|_{2,1})^ {\frac{1}{1-r}}}}
	\end{equation}
	\begin{equation} \label{NewUpdateWeight5}
		\gamma^{(v)} = \frac{( \| \mathbf{S}-\mathbf{S}^{(v)}\|_{F}^{2})^ {\frac{1}{1-r}}}{\sum_{v=1}^{V}{( \| \mathbf{S}-\mathbf{S}^{(v)}\|_{F}^{2})^ {\frac{1}{1-r}}}}
	\end{equation}
	
	Although we have ignored  the non-negative constraint, we can deduce that the final solutions Eqs.~(\ref{NewUpdateWeight3}), (\ref{NewUpdateWeight4}), and (\ref{NewUpdateWeight5}) still satisfy the constraint.
	
	Algorithm~\ref{Algorithm1} summarizes the detailed optimization procedure of CONDEN-FI. In the initialization of Algorithm~\ref{Algorithm1}, $\mathbf{S}^{(v)}$ is initialized using the $k$-nearest neighbor graph in \cite{GMC}. $\mathbf{D}_{v_{1}}$ is set as an identity matrix. $\lambda^{(v)}$, $\eta^{(v)}$, and $\gamma^{(v)}$ are each set to $\frac{1}{V}$ for every view. Both $\mathbf{B}$ and $\mathbf{B}^{(v)}$ are initialized randomly.
	
	\begin{algorithm}
		\caption{Iterative algorithm of CONDEN-FI}\label{Algorithm1}
		\KwIn{Multi-view data matrices $\{\mathbf{X}^{(v)}\}_{v=1}^{V}$, and parameters $r$,  $\theta$, and $\alpha$.}
		\textbf{Initialize:} $\mathbf{S}^{(v)}$, $\mathbf{D}_{v_{1}}$,  $\mathbf{B}$, $\mathbf{B}^{(v)}$, $\lambda^{(v)}$, $\eta^{(v)}$, and $\gamma^{(v)}$.\\
		\Begin
		{
			\While{not convergent}{
				\quad Update $\mathbf{W}^{(v)}$ according to (\ref{NewupdateWv1}) or (\ref{NewUpdateWv2}); \\
				\quad Update $\mathbf{B}$ according to (\ref{NewupdateB}); \\
				\quad Update $\mathbf{B}^{(v)}$ according to (\ref{NewUpdateBv3}); \\
				\quad Update $\mathbf{S}$ according to (\ref{NewUpdateS3}); \\
				\quad Update $\lambda^{(v)}$, $\eta^{(v)}$, and $\gamma^{(v)}$ within $\boldsymbol{\Psi}^{(v)}$ \\
				\quad using (\ref{NewUpdateWeight3}), (\ref{NewUpdateWeight4}), and (\ref{NewUpdateWeight5}), respectively. \\
			}
			Rank the features in descending order based on the $\ell_{2}$-norm of the rows of $\{\mathbf{W}^{(v)}\}_{v=1}^{V}$.\\
			Rank the instances in descending order according to the measure $MvIS$.\\
			\KwOut{Select the top  $l$  feaures and  $m$ instances.}
		}
	\end{algorithm}
	
	\subsection{Time Complexity and Convergence Analysis}\label{section3.4} 
	As observed in the procedure of CONDEN-FI in Algorithm 1, the variables $\mathbf{W}^{(v)}$, $\mathbf{B}$,  $\mathbf{B}^{(v)}$, $\mathbf{S}$, and $\boldsymbol{\Psi}^{(v)}$  are updated alternately. For updating $\mathbf{W}^{(v)}$, the main computation involves calculating the inverse of an $n \times n$ matrix in Eq.(\ref{NewUpdateWv2}) if $d_{v} > n$, or a $d_{v} \times d_{v}$ matrix in Eq.(\ref{NewupdateWv1}) otherwise, with a computational complexity of $\mathcal{O}(\min(d_{v}^{3}, n^{3}) + cnd_{v})$. Similar to update $\mathbf{W}^{(v)}$, the computational complexity of updating $\mathbf{B}$ and $\mathbf{B}^{(v)}$  is both $\mathcal{O}(n^{3})$. For updating $\mathbf{S}$, the primary computational cost comes from the matrix multiplication operation, which has a complexity of $\mathcal{O}(cn)$. For updating $\lambda^{(v)}$, $\eta^{(v)}$, and $\gamma^{(v)}$ in $\boldsymbol{\Psi}^{(v)}$, only element-based operations are involved, making the computational complexity negligible. Hence, the total main time 
	complexity of Algorithm 1 is $\mathcal{O}(Vn^{3}+cn\sum_{v=1}^{V}{d_{v}})$ for each iteration.

	Since the objective function in Eq.~(\ref{NewObjFunc}) is not jointly convex with respect to all variables $\mathbf{W}^{(v)}$, $\mathbf{B}$, $\mathbf{B}^{(v)}$, $\mathbf{S}$, and $\boldsymbol{\Psi}^{(v)}$, we divide the optimization problem (\ref{NewObjFunc})  into five sub-problems in Algorithm 1. Regarding solving the sub-problems (\ref{NewUpdateWv}), (\ref{NewUpdateB}), (\ref{NewUpdateBv}), and (\ref{NewUpdateWeight}) w.r.t. the variables $\mathbf{W}^{(v)}$, $\mathbf{B}$, $\mathbf{B}^{(v)}$, and $\boldsymbol{\Psi}^{(v)}$, respectively, all have closed-form solutions, which can guarantee the convergence of updating these variables according to~\cite{FSASL} and \cite{UEAF}.  Regarding solving the sub-problem (\ref{NewUpdateS}) w.r.t. $\mathbf{S}$, the convergence of updating $\mathbf{S}$ can be achieved as shown in \cite{GMC}. Hence, the convergence of Algorithm 1 can be guaranteed based on the updating rules of these sub-problems. Moreover, the experimental section will provide an empirical validation of the convergence behavior of Algorithm 1.

	\section{Experiments} \label{Experiments}
	In this section, extensive experiments are conducted on some real-world datasets to validate the effectiveness of the proposed method CONDEN-FI through comparisons with several state-of-the-art methods for feature and instance co-selection.
	
	\subsection{Datasets}
	We conduct experiments on eight public multi-view datasets, including three image datasets:  MSRC\_V1\footnote{http://research.microsoft.com/en-us/projects/objectclassrecognition/}, YaleB\footnote{http://vision.ucsd.edu/leekc/ExtYaleDatabase/ExtYaleB.html}, and Usps\footnote{https://www.csie.ntu.edu.tw/~cjlin/libsvmtools/datasets/}; two document  datasets: BBCSport\footnote{http://mlg.ucd.ie/datasets/segment.html} and NGs\footnote{http://lig-membres.imag.fr/grimal/data.html}; a hypertext dataset: WebKB\footnote{http://www.cs.cmu.edu/~webkb/}; a scientific article citation dataset: Cora$^{5}$, and a vehicle trace dataset: Sensit Vehicle$^{3}$. The detailed statistics of these multi-view datasets are provided in Table~\ref{DatasetDetail_table}.

	\begin{table}[htbp]
		\tabcolsep 0pt
		\caption{Details of datasets}\label{DatasetDetail_table}
		\vspace*{-10pt}
		\renewcommand\tabcolsep{1.5pt} 
		\begin{flushleft}
			\def\temptablewidth{\textwidth}
			\resizebox{0.5\temptablewidth}{!}{
				\begin{tabular}{@{\extracolsep{\fill}}lccccc}
					\toprule
					Datasets & Abbr. & Views & Instances & Features & Classes \\
					\hline
					MSRC\_V1 & MSRC & 6 & 210 & 1302/48/512/100/256/210 & 7 \\
					YaleB & YaleB & 3 & 650 & 2500/3304/6750 & 10 \\
					BBCSport & BBCS & 4 & 116 & 1991/2063/2113/2158 & 5 \\
					NGs & NGs & 3 & 500 & 2000/2000/2000 & 5 \\
					WebKB & WebKB & 2 & 1051 & 2949/334 & 2 \\
					Cora & Cora & 2 & 2708 & 1432/2223 & 7 \\ 
					Usps & Usps & 2&9298&256/32&10\\
					Sensit Vehicle & Sensit & 2 & 10200 & 50/50 & 3 \\
					\hline
			\end{tabular}}
		\end{flushleft}
	\end{table}

	\begin{table*}[htbp]
		\centering
		\caption{Classification performance of different methods on eight datasets with  selection of 20\%  instances and 30\% features.}
		\label{Ins2Fea3 ACC_F1 table}
		\vspace*{-5pt}
		\def\temptablewidth{\textwidth}
		{\rule{\temptablewidth}{0.2pt}}
		\begin{tabular*}{\temptablewidth}{@{\extracolsep{\fill}}cccccccccccc}
			
			\toprule
			Metrics & Datasets & CONDEN-FI & UFI & DFIS & sCOs & sCOs2 & MAMD & CCSD & CDMN & MFSC & MSCC \\ 
			\midrule
			\multirow{8}{*}{ACC} 
			& MSRC  &  $\mathbf{0.7952}$ & $0.7202$ & $0.5893$ & $0.2441$ & $0.5595$ & $0.5536$ & $0.4893$ & $0.3488$ & $0.6686$ & $0.6302$ \\ 
			& YaleB   & $\mathbf{0.9131}$ & $0.7077$ & $0.6262$ & $0.4019$ & $0.7231$ & $0.3519$ & $0.3815$ & $0.6108$ & $0.7689$ & $0.7595$ \\ 
			& BBCS  & $\mathbf{0.4430}$ & $0.3118$ & $0.2323$ & $0.3011$ & $0.3011$ & $0.3355$ & $0.3161$ & $0.1850$ & $0.2396$ & $0.2500$  \\ 
			& NGs  &  $\mathbf{0.6010}$ & $0.4750$ & $0.5210$ & $0.4400$ & $0.3950$ & $0.4850$ & $0.3490$ & $0.5080$ & $0.3005$ & $0.4044$ \\
			& WebKB  & $\mathbf{0.9436}$ & $0.7872$ & $0.9106$ & $0.8918$ & $0.9025$ & $0.8252$ & $0.9103$ & $0.9096$ & $0.8007$ & $0.9257$ \\ 
			& Cora   & $\mathbf{0.5067}$ & $0.3041$ & $0.4234$ & $0.3147$ & $0.3420$ & $0.3470$ & $0.4559$ & $0.4592$ & $0.3260$ & $0.3914$ \\
			& Usps    & $\mathbf{0.8601}$ & $0.8371$ & $0.7567$ & $0.5872$ & $0.7883$ & $0.7270$ & $0.6760$ & $0.5311$ & $0.6802$ & $0.8357$  \\ 
			& Sensit  & $\mathbf{0.7573}$ & $0.5234$ & $0.5724$ & $0.6358$ & $0.6951$ & $0.6034$ & $0.6556$ & $0.5575$ & $0.6395$ & $0.7018$ \\ 
			\midrule
			\multirow{8}{*}{F1} 
			& MSRC  &  $\mathbf{0.7981}$ & $0.7228$ & $0.5867$ & $0.2228$ & $0.5402$ & $0.5284$ & $0.4915$ & $0.2512$ & $0.6478$ & $0.6117$ \\ 
			& YaleB  &  $\mathbf{0.9152}$ & $0.7208$ & $0.6392$ & $0.4629$ & $0.7360$ & $0.3735$ & $0.3948$ & $0.6019$ & $0.7732$ & $0.7643$ \\
			& BBCS  & $\mathbf{0.2451}$ & $0.0951$ & $0.1082$ & $0.1056$ & $0.1061$ & $0.1394$ & $0.1149$ & $0.0985$ & $0.0773$ & $0.0800$ \\ 
			& NGs   & $\mathbf{0.5946}$ & $0.4037$ & $0.5207$ & $0.4544$ & $0.3386$ & $0.4512$ & $0.3180$ & $0.4735$ & $0.2347$ & $0.3882$ \\
			& WebKB  & $\mathbf{0.9130}$ & $0.4405$ & $0.8514$ & $0.7796$ & $0.8213$ & $0.6687$ & $0.8522$ & $0.8575$ & $0.6106$ & $0.8740$ \\
			& Cora  &  $\mathbf{0.4556}$ & $0.0666$ & $0.3663$ & $0.1375$ & $0.2395$ & $0.2490$ & $0.4127$ & $0.4040$ & $0.1523$ & $0.3045$ \\
			& Usps  &  $\mathbf{0.8454}$ & $0.8303$ & $0.7483$ & $0.5748$ & $0.7735$ & $0.7205$ & $0.6537$ & $0.5298$ & $0.6513$ & $0.8261$ \\
			& Sensit  & $\mathbf{0.7561}$ & $0.4430$ & $0.5457$ & $0.6400$ & $0.7016$ & $0.5784$ & $0.6484$ & $0.5561$ & $0.6347$ & $0.6994$ \\
			\bottomrule
		\end{tabular*}
	\end{table*}

	\subsection{Experimental Setup}
	Similar to the previous unsupervised  feature and instance co-selection methods \cite{UFI, DFIS}, we use a co-selection method to identify the most informative instances and features from each dataset. The entire dataset is then represented using the selected features. Subsequently, an SVM classifier is trained on the selected instances and their corresponding labels, and then used to predict the labels of the unselected instances. Two commonly used metrics, classification accuracy (ACC) and F1 score (F1), are employed to evaluate the quality of instances and feature subsets selected by various co-selection algorithms. Let $y_i$ and $\hat{y}_i$ represent the true label and the predicted label of the $i$-th instance, respectively. Then, ACC can be defined as follows:
	\begin{equation}
		\text{ACC}=\frac{\sum_{i=1}^{n} \mathbf{1}(\hat{y}_i = y_i)}{n},
	\end{equation}
	where $\mathbf{1}(\hat{y}_i = y_i)$ is 1 if $\hat{y}_i = y_i$, and 0 otherwise.
	Besides, the F1 score is defined as:
	\begin{equation}
		\text{F1}=\frac{1}{c} \sum_{j=1}^{c}  \frac{2 P_j  R_j}{P_j + R_j},
	\end{equation}
	where $c$ denotes the number of classes, and $P_j$ and $R_j$ represent the precision and recall of the $j$-th class, respectively. For these two metrics, larger values indicate better performance.
	
	Meanwhile, we compare the proposed method CONDEN-FI with several  state-of-the-art (SOTA)   feature and instance  co-selection methods, including  UFI~\cite{UFI}, DFIS~\cite{DFIS}, sCOs~\cite{sCOs} and sCOs2~\cite{sCOs2}. In addition, some competitive multi-view feature selection methods and instance selection methods are combined to achieve the simultaneous selection of instances and features. These baseline methods are briefly introduced as follows:

	$\bullet$ \textbf{UFI}: Unified criterion for feature and instance selection, which introduces a greedy algorithm to minimize the trace of the covariance matrix for selecting the most informative features and instances.
	
	$\bullet$ \textbf{DFIS}: It attempts to simultaneously reconstruct the latent data space using the selected features and instances while preserving the global structure of the data.
	
	$\bullet$ \textbf{sCOs}: The method introduces a similarity preservation with $\ell_{2,1}$-norm regularization for the selection of both features and instances.

	$\bullet$ \textbf{sCOs2}: Building on sCOs, this method employs the  $\ell_{2,1-2}$-norm  to enhance sparsity and remove unimportant features and samples in the selection of the representative subset.

	$\bullet$ \textbf{MAMD}: This method combines a multi-view unsupervised feature selection method (MAMFS\cite{MAMFS}) that uses adaptive graph learning, and an unsupervised instance selection method (D-SMRS\cite{D-SMRS}) that employs self-representation learning to recursively select representative instances.
	
	$\bullet$ \textbf{CCSD}: This combined method selects features by utilizing a consensus cluster structure-based multi-view unsupervised feature selection method (CCSFS~\cite{CCSFS}), and selects informative instances using D-SMRS, similar to MAMD.
	
	$\bullet$ \textbf{CDMN}: It combines  multi-view unsupervised feature selection with consensus partition and diverse graph (CDMvFS~\cite{CDMvFS}) and unsupervised instance selection via conjectural hyperrectangles (NIS~\cite{NIS}) to simultaneously select features and instances.
	
	$\bullet$ \textbf{MFSC}: This method integrates both a multi-view feature selection method that learns a shared optimal similarity graph across all views (MFSGL~\cite{MFSGL}), and an instance selection method that uses cluster centers and boundaries to identify informative instances (CIS~\cite{CIS}).
	
	$\bullet$ \textbf{MSCC}: The combined method selects features by fusing multi-view subspace clustering with a self-representation method to preserve the data block structure in each view (MSCUFS~\cite{MSCUFS}), and selects instances using CIS, similar to MFSC.

	Since these methods UFI, DFIS, sCOs, and sCOs2 are designed for single-view data, we first combine all features from different views into a single dataset. We then use these methods to select features and samples simultaneously from the integrated dataset. The parameters of all compared methods are tuned according to the ranges given in the corresponding papers to obtain optimal results. Meanwhile, for the parameters of our method, $\alpha$ and $\theta$ are tuned within the range of \{$10^{-3}, 10^{-2}, \cdots,10^{3}$\}, and $r$ is tuned within the range of \{$2, 3, 4, 5, 6$\}. Due to the difficulty in determining the optimal number of features and instances for each dataset~\cite{UFI, sCOs2, MUFSsurvey}, we respectively vary the percentages of selected features and instances from 10\% to 50\% in increments of 10\% across all datasets and report the averaged classification results for each method. Each experiment is conducted five times independently, and the average result is recorded for subsequent comparison.

	\subsection{Experimental Results}
	In this section, we compare the classification performance of the proposed method CONDEN-FI with other SOTA  feature and instance  co-selection methods on eight benchmark datasets.  Table~\ref{Ins2Fea3 ACC_F1 table} presents the classification results  ACC and F1 of different methods when the percentages of selected features and instances are set to 30\% and 20\%, respectively. In this table, the best performance is highlighted in bold. From this table, we can see that our method CONDEN-FI is consistently better than other compared methods. As to YaleB and BBCS datasets, CONDEN-FI can improve classification performance by more than 10\% in ACC and F1 compared to the second best method. For  MSRC and NGs datasets, it achieves over a 7\% improvement in both metrics. As to Sensit dataset, CONDEN-FI gains over a 5\% improvement in both ACC and F1 in comparison with the runner-up method. For Cora dataset, CONDEN-FI outperforms the second best method by nearly 5\% in terms of ACC and F1. As to WebKB and Usps datasets, CONDEN-FI remains superior to the second best method, with improvements in both ACC and F1 approaching 2\%. In addition, CONDEN-FI surpasses all single-view based feature and instance co-selection methods, achieving an average improvement of nearly 10\% in ACC and F1 across most datasets. This demonstrates the effectiveness of the proposed method, which utilizes inter-view consistent and view-specific information from multiple views, compared to single-view methods that merely stack features from different views.
	
\begin{figure*}[!htbp]  
		\centering 
		\includegraphics[width=1\textwidth]{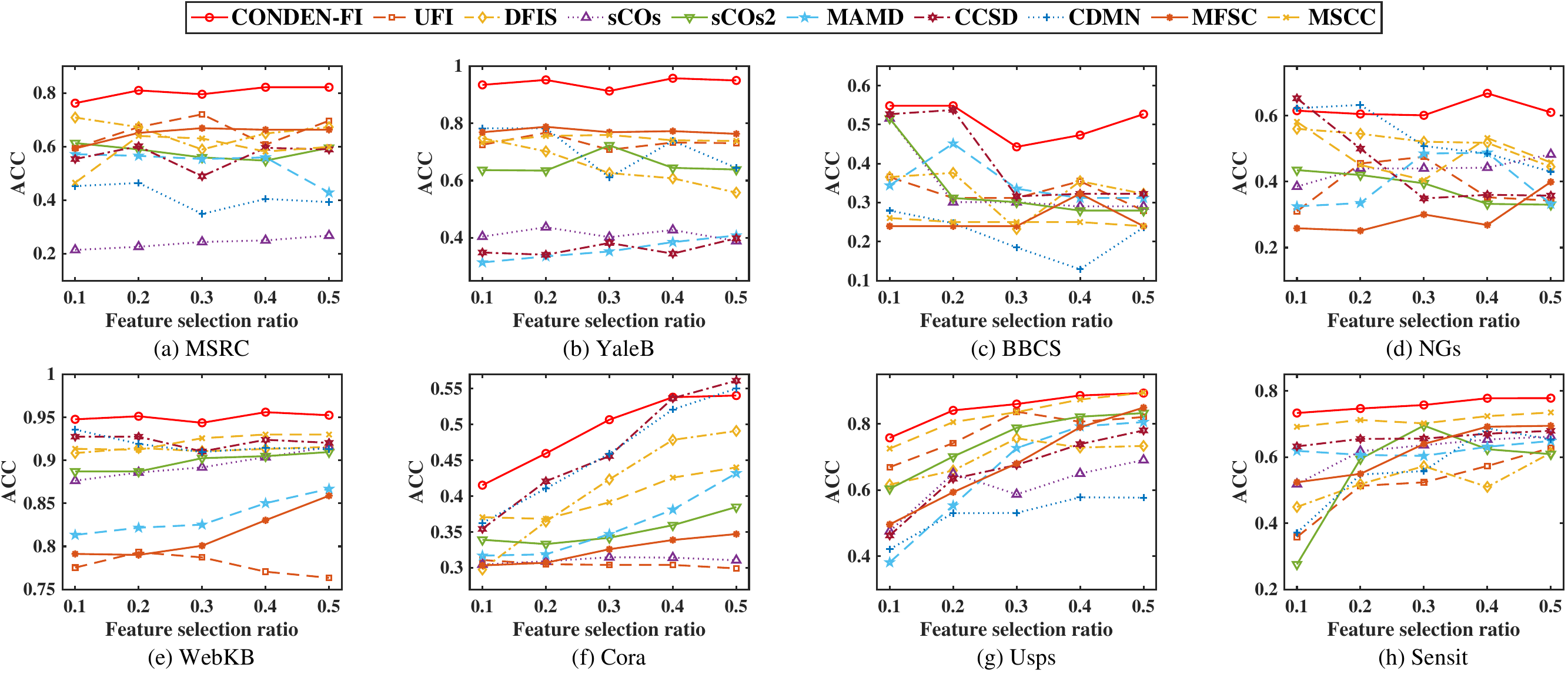}
		\caption{ACC of different methods with different ratios of selected features.}
		\label{Ins2_ACC_figure}
	\end{figure*}

\begin{figure*}[!htbp]  
		\centering 
		\includegraphics[width=1\textwidth]{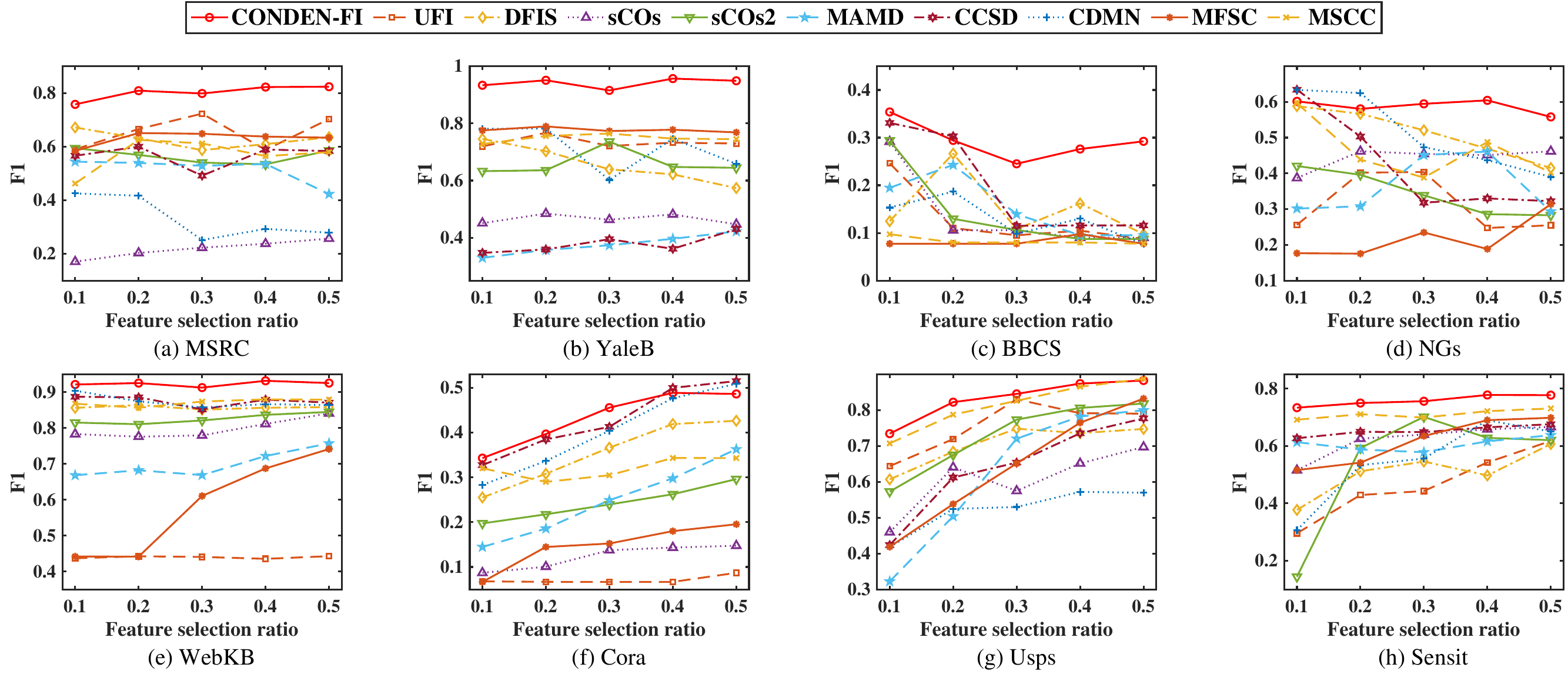}
		\caption{F1 of different methods with different ratios of selected features.}
		\label{Ins2_F1_figure}
	\end{figure*}

\begin{figure*}[!htbp]  
		\centering 
		\includegraphics[width=1\textwidth]{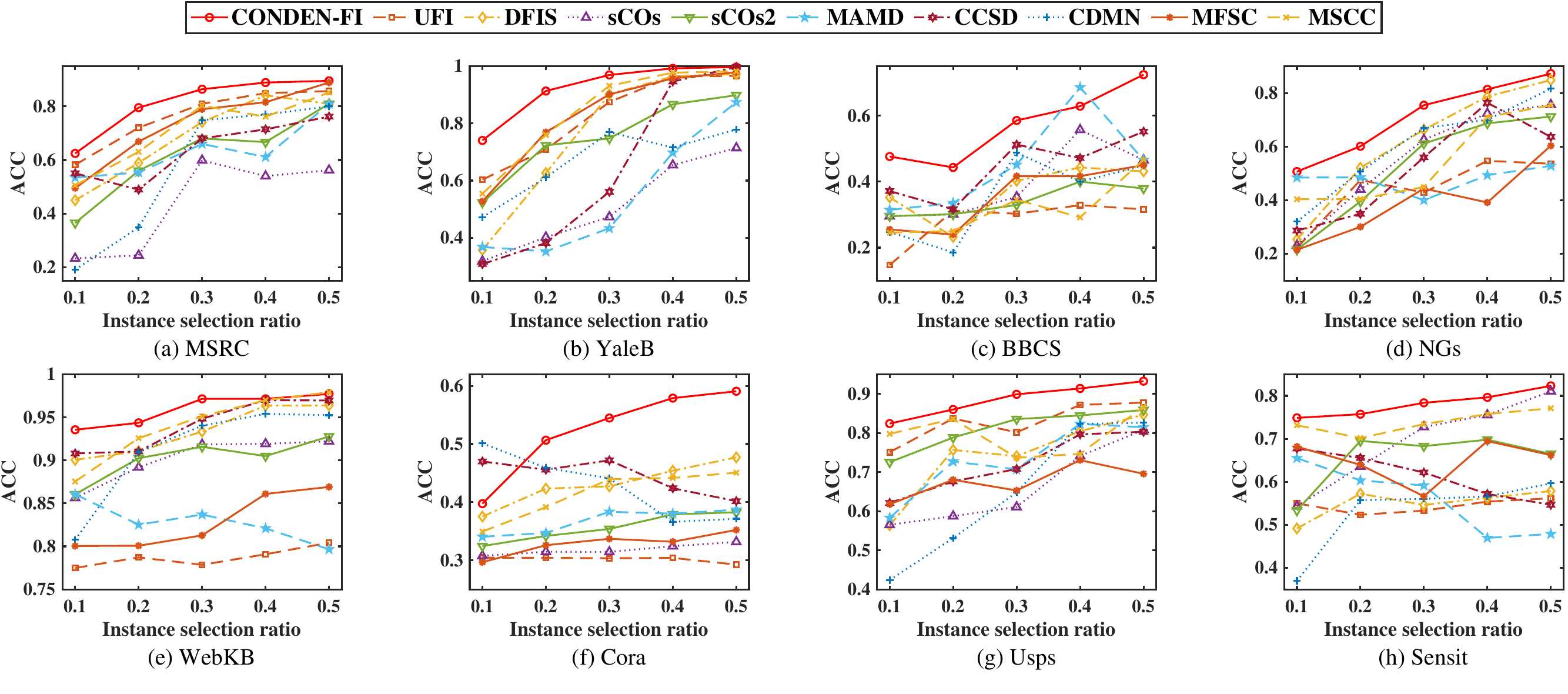}
		\caption{ACC of different methods with different ratios of selected instances.}
		\label{Fea3_ACC_figure}
	\end{figure*}

\begin{figure*}[!htbp]  
		\centering 
		\includegraphics[width=1\textwidth]{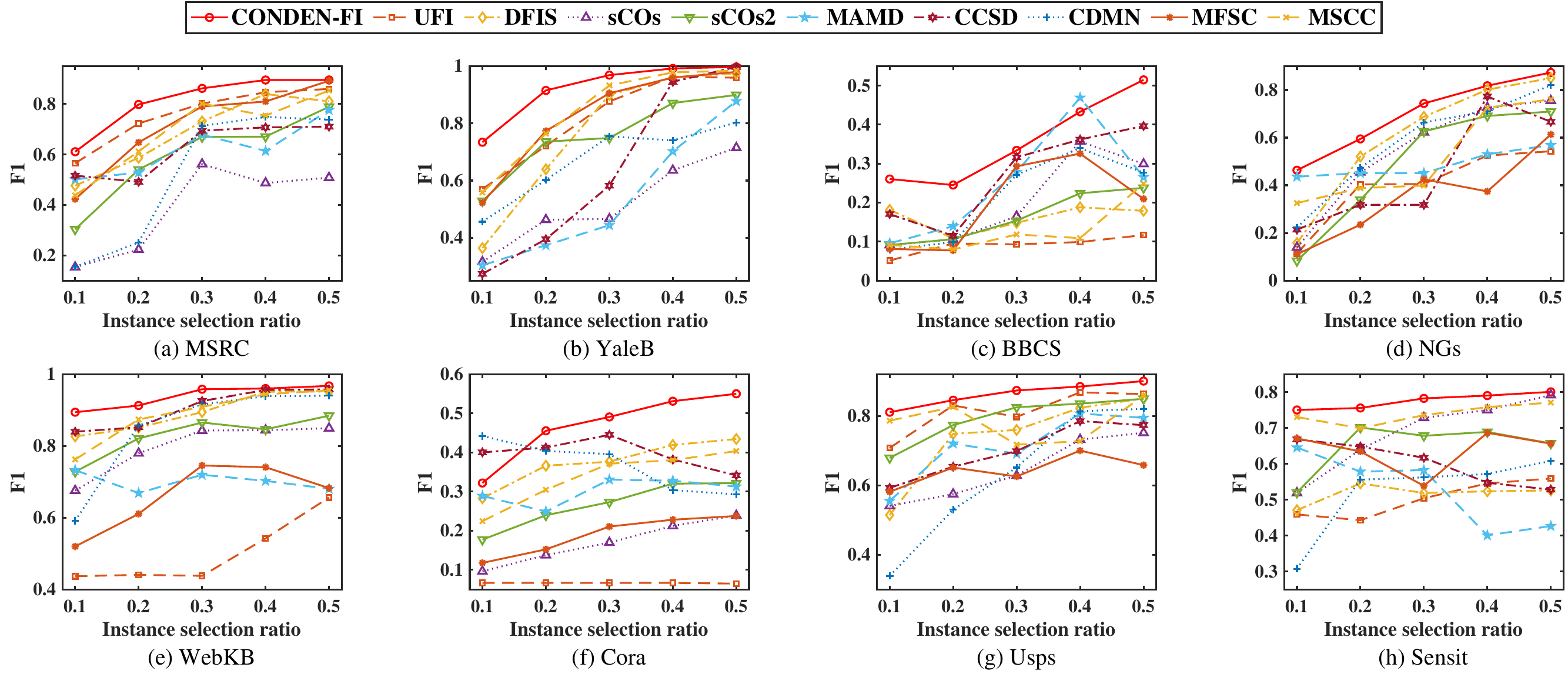}
		\caption{F1 of different methods with different ratios of selected instances.}
		\label{Fea3_F1_figure}
	\end{figure*}

	Since determining the optimal number of selected features and instances for each dataset is difficult, we also present the classification performance of different methods across  varying numbers of selected features and instances. Figs.~\ref{Ins2_ACC_figure} and \ref{Ins2_F1_figure} respectively show the ACC and F1 values of different methods as the ratio of selected features varies from 10\% to 50\%, with the ratio of selected instances fixed at 20\%. From these two figures, we can see that CONDEN-FI performs better than the other baseline methods in most cases when feature selection ratios range from  10\% to 50\%. Moreover, Figs.~\ref{Fea3_ACC_figure} and \ref{Fea3_F1_figure} show the ACC and F1 values of different methods across eight datasets, with the ratio of selected instances ranging from 10\% to 50\% and the ratio of selected features set to 30\%, respectively.  As can be seen from the results, our method CONDEN-FI outperforms other methods in most cases over a range of selected instances. The superior performance of the proposed CONDEN-FI is attributed to its ability to adaptively learn inter-view consistency and view-specific information from multi-view data, integrating these into a data self-representation framework to simultaneously select diverse instances and informative features.

	\subsection{Parameter Sensitivity}
\begin{figure*}[!htbp]  
		\centering 
		\includegraphics[width=\textwidth]{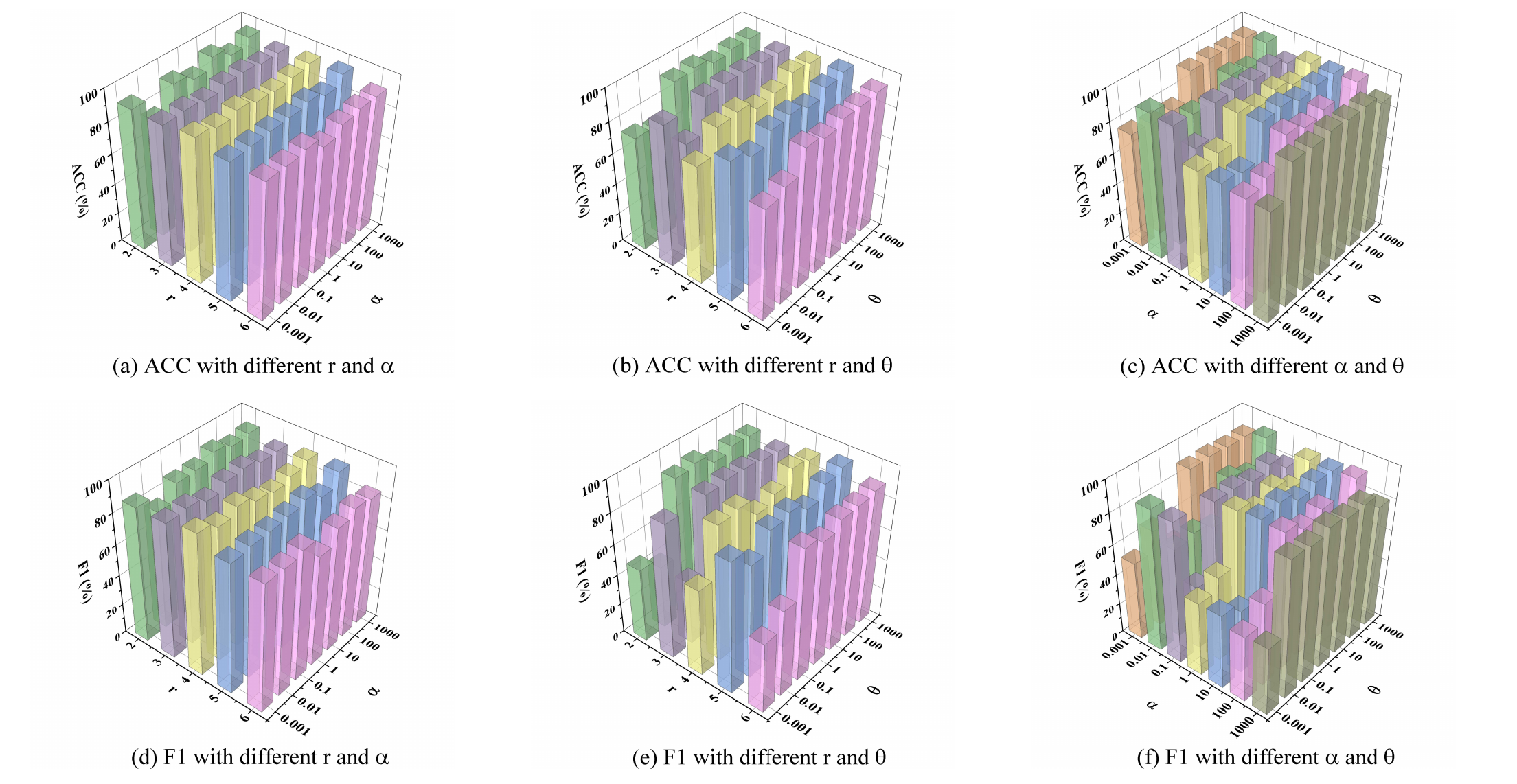}
		\caption{Performance variations with different parameters on  WebKB dataset. The first and second rows respectively show the ACC and F1 with varying parameters.}
		\label{sensitivity_webkb_figure}
	\end{figure*}

	There are three parameters, $\theta$, $\alpha$,  and $r$, in the proposed method as formulated in Eq.~(\ref{NewObjFunc}). In this section, we conduct experiments to investigate the performance variations with regards to these three parameters. Due to the page limitation, we only report the ACC and F1 results on  WebKB dataset with the feature and instance ratios fixed at 30\% and 20\%, respectively. Fig.~\ref{sensitivity_webkb_figure} illustrates the ACC and F1 values as two of the three parameters, $\alpha$, $\theta$, and $r$, vary while the third remains fixed. From this figure, we can see that when $\theta$ is fixed and $\alpha$ and $r$ vary, the ACC and F1 values show slight fluctuations, whereas when  $\alpha$ or $r$ is fixed and $\theta$ varies, these values exhibit relatively large fluctuations. Furthermore, our method achieves relatively superior performance in most cases with parameter combinations of $\theta$=0.1, 1 or 10, $r=2$, and $\alpha=0.001$. Hence, we can empirically fine-tune $\theta$, $\alpha$, and $r$ using the discussed smaller range.

	\subsection{Convergence Analysis}
\begin{figure}[!htbp]    
	\centering 
	\includegraphics[width=0.5\textwidth]{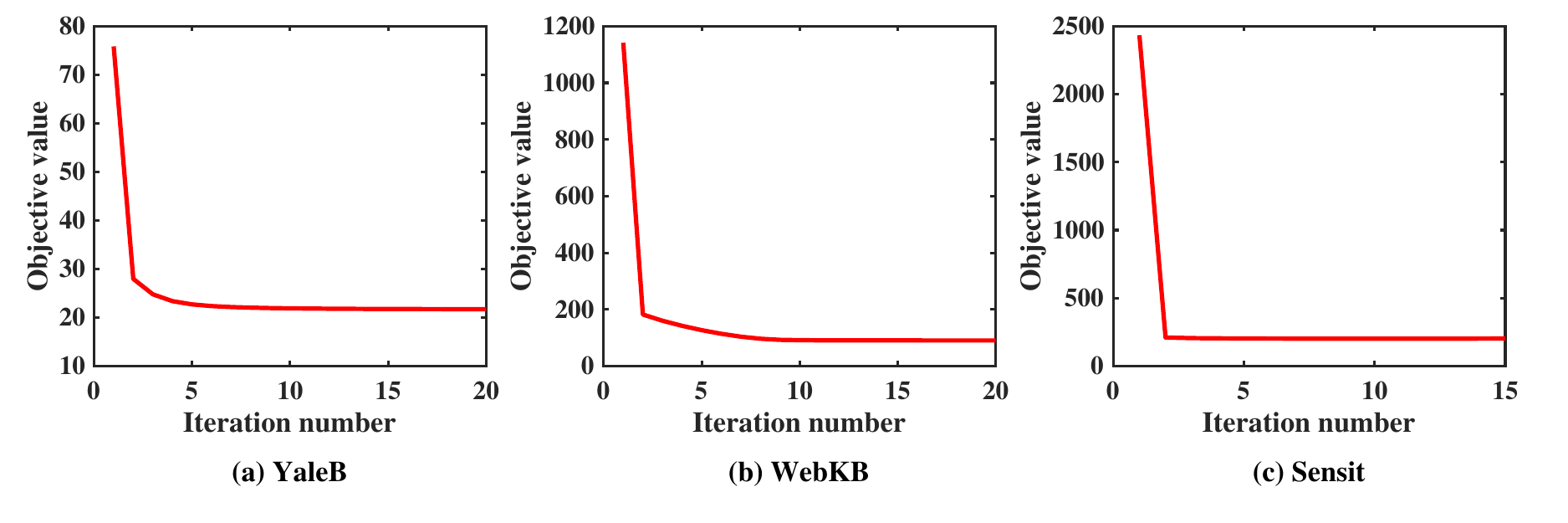}
	\caption{Convergence curves of CONDEN-FI on YaleB, WebKB and Sensit datasets.}
	\label{convergence_figure}
\end{figure}

	In Section~\ref{section3.4}, we have provided a theoretical analysis of the convergence of the proposed CONDEN-FI. In this section, we perform experiments to empirically validate its convergence. Fig.~\ref{convergence_figure} shows the changes of  the objective function values with different number of iterations on YaleB, WebKB, and Sensit datasets. As shown in this figure, the convergence curves of CONDEN-FI drop sharply within a few iterations and stabilize in nearly 20 iterations. This demonstrates that CONDEN-FI is capable of converging effectively.

\begin{table*}[!htbp]  
	\centering
	\caption{Performance comparison between CONDEN-FI and its two variants on eight datasets in terms of ACC and F1.}
	\label{ablation results}
	\vspace*{-5pt}
	\def\temptablewidth{\textwidth}
	{\rule{\temptablewidth}{0.2pt}}
	\begin{tabular*}{\temptablewidth}{@{\extracolsep{\fill}}cccccccccc}
		\toprule
		\multirow{2}{*}{Metrics} & \multirow{2}{*}{Methods} & \multicolumn{8}{c}{Datasets} \\ \cmidrule(lr){3-10}
		&                         & MSRC  & YaleB  & BBCS   & NGs    & WebKB  & Cora   & Usps   & Sensit \\ \midrule
		\multirow{3}{*}{ACC}     & CONDEN-FI                    & $\mathbf{0.7952}$ & $\mathbf{0.9131}$  & $\mathbf{0.4430}$ & $\mathbf{0.6010}$  & $\mathbf{0.9436}$  & $\mathbf{0.5067}$  & $\mathbf{0.8601}$  & $\mathbf{0.7573}$ \\ 
		& CONDEN-FI-$\mathrm{I}$             & $0.6095$ & $0.8577$  & $0.3355$ & $0.5250$  & $0.8825$  & $0.4122$  & $0.8094$  & $0.5562$ \\ 
		& CONDEN-FI-$\mathrm{II}$              & $0.5679$ & $0.8439$ & $0.3269$ & $0.4625$  & $0.8747$ & $0.3727$ & $0.5486$ & $0.5216$ \\ \midrule
		\multirow{3}{*}{F1}      & CONDEN-FI                     & $\mathbf{0.7981}$ & $\mathbf{0.9152}$  & $\mathbf{0.2451}$  & $\mathbf{0.5946}$  & $\mathbf{0.9130}$  & $\mathbf{0.4556}$  & $\mathbf{0.8454}$  & $\mathbf{0.7561}$ \\ 
		& CONDEN-FI-$\mathrm{I}$             & $0.6173$  & $0.8601$  & $0.1575$  & $0.5078$  & $0.8013$  & $0.3257$  & $0.8063$  & $0.5304$ \\ 
		& CONDEN-FI-$\mathrm{II}$               & $0.5384$ & $0.8458$ & $0.0984$ & $0.4466$ & $0.7747$ & $0.3019$ & $0.5053$ & $0.5130$ \\ \bottomrule
	\end{tabular*}
\end{table*}

	\subsection{Ablation Study}
	In this section, we conduct ablation experiments to demonstrate the effectiveness of the component terms in the proposed CONDEN-FI, as described in Eq.~(\ref{NewObjFunc}). To this end, two variant methods of CONDEN-FI are developed as follows:
	
	(1) CONDEN-FI-$\mathrm{I}$: It performs feature and instance co-selection without adaptively learning the inter-view consistency similarity matrix used for selecting diverse samples.
	\begin{equation} \label{NewVariant1}
		\begin{aligned}
			\min _{\Omega_{1}} &\sum_{v=1}^{V}{(\| {{\mathbf{W}^{(v)}}^T }{\mathbf{X}^{(v)}}-{{\mathbf{W}^{(v)}}^T }{\mathbf{X}^{(v)}}( \mathbf{B}+{\mathbf{B}^{(v)}} ) \|_{F}^{2}  }\\
			&+{{\eta^{(v)}}^{r}{{\| {\mathbf{B}^{(v)}} \|}_{2,1}}+\lambda^{(v)}}^{r}{{\| {\mathbf{W}^{(v)}} \|}_{2,1}})
			+\theta {{\| \mathbf{B} \|}_{2,1}},\\
		\end{aligned}
	\end{equation}
	where $\Omega_{1}=\{\mathbf{W}^{(v)}, \mathbf{B}, \mathbf{B}^{(v)}, \eta^{(v)}, \lambda^{(v)}\}$.
	
	(2) CONDEN-FI-$\mathrm{II}$: It learns view-specific information for simultaneous selection of features and instances while ignoring the shared information among multi-view data.
	\begin{equation} \label{Variant-2}
		\begin{aligned}
			\min _{\Omega_{2}} &\sum_{v=1}^{V} (
			{|| {{\mathbf{W}^{(v)}}^T }{\mathbf{X}^{(v)}}-{{\mathbf{W}^{(v)}}^T }{\mathbf{X}^{(v)}} \mathbf{B}^{(v)} ||_{F}^{2}  } 
			+{\eta^{(v)}}^{r}{{||  {\mathbf{B}^{(v)}} || }_{2,1}}\\
			&+{\lambda^{(v)}}^{r}{{||  {\mathbf{W}^{(v)}} || }_{2,1}}   + \frac{\alpha}{2}\sum_{i,j=1}^{n}{
				{||  \mathbf{B}^{(v)}_{i\cdot} - \mathbf{B}^{(v)}_{j\cdot} ||_{2}^{2} s_{ij}^{(v)}}}),\\
		\end{aligned}
	\end{equation}
	where $\Omega_{2}=\{\mathbf{W}^{(v)}, \mathbf{B}^{(v)}, \eta^{(v)}, \lambda^{(v)}\}$.
	
	The constraints of CONDEN-FI-$\mathrm{I}$ and CONDEN-FI-$\mathrm{II}$ regarding the optimization variables are the same as those in Eq.~(\ref{NewObjFunc}).
	
	Table~\ref{ablation results} presents the comparison results of CONDEN-FI and its two variants on eight datasets. From this table, we can see that CONDEN-FI achieves the best performance across all datasets in terms of ACC and F1.  Compared to the variant CONDEN-FI-$\mathrm{I}$, CONDEN-FI shows a substantial enhancement in classification performance. This demonstrates the effectiveness of adaptive learning of the inter-view consistency similarity matrix, which can facilitate the selection of diverse samples and capture the common local manifold structure inherent in multi-view data. Besides, CONDEN-FI significantly outperforms CONDEN-FI-$\mathrm{II}$ across all datasets, which demonstrates that learning the reconstruction of the shared representation across different views is beneficial for improving performance. Hence, CONDEN-FI integrates the learning of the inter-view consistency similarity matrix with the shared representation across different views, enabling them to enhance each other and support the selection of diverse instances and informative features, which contributes to improving classification performance.

	\section{Conclusion}  \label{Conclusion}
	In this paper, we proposed a novel multi-view unsupervised feature and instance co-selection method CONDEN-FI by utilization of the shared and view-specific information among different views. CONDEN-FI jointly selects features and instances to reconstruct a reduced-dimensional data space by learning inter-view consistency and view-specific representations. Meanwhile, the proposed method is designed to adaptively learn a view-consensus similarity graph through the integration of view-specific similarity graphs, allowing for diverse sample selection in multi-view data.  Extensive experimental results on real-world multi-view datasets demonstrated the superiority of the proposed method in comparison with several state-of-the-art methods.

	\section*{Acknowledgments}
	This work was supported by the National Natural Science Foundation of China (No. 72495122), the Natural Science Foundation Project of Sichuan Province (No. 2024NSFSC0504), and the Youth Fund Project of Humanities and Social Science Research of Ministry of Education (No. 21YJCZH045).



	
	%

\end{document}